\def\year{2018}\relax
\documentclass[letterpaper]{article} 
\usepackage{aaai18}  
\usepackage{times}  
\usepackage{helvet}  
\usepackage{url}  
\usepackage{graphicx}  
\usepackage{color}
\usepackage{bm}
\usepackage{amsthm}
\usepackage{comment}
\usepackage{booktabs}
\usepackage{amsmath}
\usepackage{amssymb}
\usepackage{wrapfig}
\usepackage{multirow}
\usepackage{caption}
\usepackage[ruled]{algorithm2e}

\newtheorem{prop}{Proposition}

\newcommand{\xiaocheng}[1]{{{\color{blue}{[#1]}}}}

\begin{document}
%
\title{Convolution on Graph: A High-Order and Adaptive Approach}

\author{Zhenpeng Zhou \and Xiaocheng Li\\
\texttt{\{zhenpeng, chengli1\}@stanford.edu}\\}
\maketitle
\begin{abstract}
In this paper, we presented a novel convolutional neural network framework for graph modeling, with the introduction of two new modules specially designed for graph-structured data: the $k$-th order convolution operator and the adaptive filtering module. Importantly, our framework of High-order and Adaptive Graph Convolutional Network (HA-GCN) is a general-purposed architecture that fits various applications on both node and graph centrics, as well as graph generative models. We conducted extensive experiments on demonstrating the advantages of our framework. Particularly, our HA-GCN outperforms the state-of-the-art models on node classification and molecule property prediction tasks. It also generates 32\% more real molecules on the molecule generation task, both of which will significantly benefit real-world applications such as material design and drug screening.
\end{abstract}

\section{Introduction}

Convolutional neural networks (CNNs) have achieved great success in various tasks from computer vision \cite{Huang:2016wa}, speech recognition \cite{Zhang:2017up} and natural language processing \cite{Conneau:2017to}. CNN provides us an efficient and effective architecture to learn meaningful representations for graphics and texts. In recent years, researchers thrive to extend the operator of convolution and develop CNN architectures for graphs, which possibly have more complicated structures than images. The graph convolutional networks are usually applied on the following two centrics of learning tasks:
\begin{itemize}
\item Node centric: the prediction tasks related to the nodes in a graph. The graph convolutional networks usually do so via outputting a feature vector for each node in the graph, which meaningfully reflects the node's property and neighborhood structure. For example, in social networks, the vectors can be used for tasks like node classification and link prediction. Sometimes, this is related to node representation learning. 
\item Graph centric: the prediction tasks related to the graphs. For example, in the context of chemistry, a molecule can be viewed as a graph with atoms as nodes and bonds as edges. graph convolutional networks are constructed to encode the molecules meaningfully in terms of their physical and chemical properties. These tasks are therefore the key to many real-world applications such as material design and drug screening. In this context, graph convolutional networks usually find a way to encode the graph and use the encodings for graph prediction tasks.
\end{itemize}

Early efforts on designing neural networks for graphs date back to the works of Gori et al. and Scarselli et al. \cite{gori2005new,scarselli2009graph}, in which they built sequential or recurrent network architectures for graph-structured data. The study of Bruna et al. \cite{bruna2013spectral}, Edwards et al. \cite{Edwards:2016vy} and Defferrard et al. \cite{defferrard2016convolutional} further developed the idea of spectral filtering/convolution which operates on the graph spectrum. Henaff et al. \cite{henaff2015deep} extended the graph convolutional networks to large scale datasets like ImageNet Object Recognition, text categorization, and bioinformatics. Meanwhile, Niepert et al. \cite{niepert2016learning} proposed an approach of PATCHY-SAN, which defined operations of node sequence selection, neighborhood assembly, and graph normalization. Atwood et al. \cite{Atwood:2016wq} presented diffusion-convolutional neural networks (DCNNs) model for graph-structured data. As we will show later, these models successfully made CNN work under the graph settings, but they still lack of careful considerations for the specialties of the graph structures in the network design.

Also there are several newly-published results on conducting graph convolutions dynamically. Jia et al. \cite{Jia:2016wt} proposed the Dynamic Filter Network, where filters are generated dynamically conditioned on the input features. Simonovsky et al. \cite{Simonovsky:2017tv} extended that idea to graphs, using edge-conditioned dynamic weights for graph convolutions. The work of Verma et al. \cite{Verma:2017tb} managed to determine the shape of filters as a function of the features in previous network layers. Manessi et al. \cite{Manessi:2017wp} proposed a model to learn temporal information from graphs that have a changing structure overtime. Li et al. \cite{Li:2017ud} proposed a general and flexible graph convolutional network (EGCN) to deal with data with diverse or undefined dimensions. All these research found one way or another to dynamically utilize the graph data, and our work can be seen as a further endeavor with the introduction of the adaptive convolution module. 

Besides, a great amount of research on graph-centric tasks concentrates on the application of molecule fingerprints. The molecule fingerprinting refers to a quantitative encoding for the molecules that can be used for molecule property summarization and prediction. Prior to the usage of CNN, the graph kernels have dominated many learning and prediction tasks for (molecule) graphs \cite{kondor2002diffusion,shervashidze2009efficient,shervashidze2011weisfeiler}. The paper of Duvenaud et al. \cite{duvenaud2015convolutional} first introduced CNN for encoding the molecules and Kearnes et al. \cite{kearnes2016molecular} further improved the results. Gilmer et al. \cite{Gilmer:2017tl} defined a Message Passing Neural Networks (MPNNs) for molecules to reformulate existing models into a single common framework with a message passing interpretation. It will be shown later that our work greatly improved the state-of-the-art performance on molecule-relevant tasks with a network architecture that better captures the properties of the molecules graphs. 

In this work, we proposed a novel graph convolutional network architecture named High-order and Adaptive Graph Convolutional Network (HA-GCN). The most related work to ours is the graph convolutional network (GCN) \cite{kipf2016semi}, in which the convolution operator only reaches one-hop neighbors. Our high-order operator provides an efficient design of convolution that reaches $k$-hop neighbors. Furthermore, we introduced an adaptive filtering module that adjusts the weights of convolution operators dynamically based on the local graph connections and node features. Compared with the work of Li et al. \cite{li2015gated} which introduced the modern idea of LSTM into graph settings, our adaptive module can be interpreted as a graph realization of the attention mechanism proposed in \cite{xu2015show}. Most importantly, unlike the previous graph networks designed for either node-centric or graph-centric task, our HA-GCN framework is general-purposed and capable of fulfilling both. Additionally, we  constructed a graph generative model with HA-GCN for the task of molecule generation, achieving a significant improvement over the state-of-the-art model.
 
Our contribution is two-fold:

\begin{itemize}
\item We introduced two new modules for graph-structured data and built a novel graph convolutional network framework of HA-GCN.
\item We developed a general-purposed architecture that can be applied for node-centric prediction, graph-centric prediction and graph generative modeling. Our architecture achieved state-of-the-art performance uniformly on all the tasks.
\end{itemize}

The rest of the paper is organized as follows: first we provide some preliminaries for the graph model and a brief discussion of several frameworks of graph convolutional networks. Then we introduce the key ideas of high-order convolution operator and adaptive filtering module. Furthermore, we present our framework of HA-GCN and several experiments to demonstrate its performance. Finally, we summarize the scope of HA-GCN applications and point out the potential future directions.

\section{Preliminaries}

\subsection{The Graph Model\footnote{In this paper, we use the terminology ``graph" to refer to the graph/network structure of data and ``network" for the architecture of machine learning models.}}
In this subsection, we provide the preliminaries and notations for the graph model. A graph $\mathcal{G}$ is denoted as a pair $(V,E)$ with $V=\{v_1,...,v_n\}$ the set of nodes (vertices) and $E\in V\times V$ the set of edges. Here we do not distinguish the undirected and directed graphs in terms of notations since our framework works for both cases. Each graph can be represented by a $n$-by-$n$ adjacency matrix $A$ where $A_{i,j}=1$ if there is an edge from $v_i$ to $v_j$ and $A_{i,j}=0$ otherwise. Based on the adjacency matrix, we can have a distance function $d(v_i,v_j)$ to represent the graph distance from $v_i$ to $v_j$ (the minimum length of paths connecting $v_i$ and $v_j$). Additionally, we assume that each node $v_i$ is associated with a feature vector $X_i\in \mathcal{R}^m$, and compactly we use $X = \left(X_1^T,X_2^T,...,X_n^T\right) \in \mathcal{R}^{n\times m}$ to denote the feature matrix.

\subsection{Graph Convolutional Networks (GCNs)}

In this subsection, we briefly review several GCN structures from previous works to provide some intuitions for the design of convolution on graph. At the first place, the convolution operator at a specific node $v_j$ in graph $\mathcal{G}$ can be generally expressed as 

$$L_{conv}(j) = \sum_{i \in \mathcal{N}_{j}} w_{ij} X_i + b_j.$$
Here $X_i\in R_m$ is the input feature for node $v_i$, $b_j$ is the bias term and $w_{ij}$ is the weight which can be non-stationary and vary with respect to $j$. The set $\mathcal{N}_{j}$ defines the scope of convolution. For traditional applications, the CNN architecture is usually designed for a low-dimensional grid with the same connection pattern for every node. For example,  images can be viewed as two-dimensional grids (for each of the RGB channels, or gray scale channel), and the underlying graph $\mathcal{G}$ is formed by connecting adjacent pixels. Then $\mathcal{N}_j$ can be simply defined as a fixed-size block or window around pixel $j$.

In the more general graph settings, one can define $\mathcal{N}_j$ as the set of nodes that are adjacent to $v_j$. 
For example, the core of the fingerprint (FP) convolution operator in the work of Duvenaud et al. \cite{duvenaud2015convolutional} is to compute the average over neighbors, i.e. $w_{ij}=1$ for all $(i,j)$. With the help of adjacency matrix $A,$ we can write the operator as 
\begin{equation}
L_{FP} = AX.
\label{FP}
\end{equation}
The multiplication of $A$ and the feature matrix $X$ results in a feature averaging over neighbor nodes. One step further, the paper of node-GCN \cite{kipf2016semi} applied linear weighting and non-linear transformation in addition to the averaging:
\begin{equation}
L_{node-GCN} = \sigma(AXW).
\label{node-GCN}
\end{equation}
The weight matrix $W$ and the function $\sigma(\cdot)$ perform a linear and non-linear transformation on the feature $X$ respectively. 

The papers of Bruna et al.\cite{bruna2013spectral} and Defferrard et al. \cite{defferrard2016convolutional} took a different approach by conducting convolution on the spectrum of a graph Laplacian. Let $H$ be the graph Laplacian and its orthogonal decomposition $H=U\Lambda U^T$ (where $U$ is a orthogonal matrix and $\Lambda $ is a diagonal matrix). Instead of appending the weight matrix as in (\ref{node-GCN}), the spectral convolution considers a parameterized convolution operator on $H.$ Precisely, 
\begin{equation}
L_{spectral} = Ug_{\theta}(\Lambda)U^TX.
\label{spectral}
\end{equation}
Here the function $g_{\theta}(\cdot)$ is a polynomial function which is element-wisely applied on the diagonal matrix $\Lambda.$ 

When discussing the advantage of spectral convolution, the authors mentioned that a $k$-order polynomial choice of $g_{\theta}(\cdot)$ is exactly  $k$-localized on graph, which means the convolution reaches as far as the $k$-hop neighbors. Compared to the one-hop neighbor averaging in (\ref{FP}) and (\ref{node-GCN}), this allows faster information propagation over the graph. However, the the choice of polynomial $g_{\theta}(\cdot)$ does not give an exact convolution operator for $k$-hop neighbors, as not all neighbors are equally weighted as assumed in convolution, with the fact in mind that $U\Lambda^nU^T=A^n$. This motivates the proposal of our high-order convolution operator. Another problem with all of those convolution operators is that they are using fixed convolution weights, which are invariant across graphs. Therefore it can hardly capture the differences between the locations where the convolution operation happens.  This motivates the design of our adaptive module, which successfully takes both the local features and the graph structures into account.

\section{High-Order and Adaptive Graph Convolutional Network (HA-GCN)}

\subsection{$K$-th Order Graph Convolution}

We begin with the definition of the $k$-hop ($k$-th order) neighborhood: $\mathcal{N}_j = \{v_i\in V | d(v_i,v_j)\le k\}$ for node $v_j$. In fact, the exact $k$-hop connectivity can be obtained by the multiplication of the adjacency matrix $A$, as formally stated in the following proposition. 

\begin{prop}
Let $A$ be the adjacency matrix of a graph $\mathcal{G}$, then the $(i,j)$ entry of its $k$-th product $A^k$ is the number of $k$-hop paths from $i$ to $j$.
\end{prop}

With this proposition, we can define a $k$-th order convolution operator as follows
\begin{equation}
\widetilde{L}^{(k)}_{gconv} = \left(W_k \circ \widetilde{A}^k\right)X+B_k,
\label{ho}
\end{equation}
where
\begin{equation}
\widetilde{A}^k = \min \{A^k+I, 1\}.
\end{equation}

Here $\circ$ and $\min$ refer to element-wise matrix product and minimum respectively. The $W_k \in \mathcal{R}^{n\times n}$ is the weight matrix while $B_k \in \mathcal{R}^{n \times m}$ is the bias matrix. The $\widetilde{A}^k$ is obtained by clipping $A^k+I$ to $1$. The addition of identity matrix $I$ to $A^k$ creates a self loop for each node in the graph. And the clipping is motivated by the fact that if the matrix of $A^k$ have elements larger than one, clipping those values to $1$ will exactly lead to the convolution of $k$-hop neighborhood. The input of the operator $\widetilde{L}^{(k)}_{gconv}$ is the adjacency matrix $A \in \{0,1\}^{n\times n}$ and feature matrix $X \in \mathcal{R}^{n\times m}$. Its output has the same dimension as $X$. As the name suggests, the convolution $\widetilde{L}^{(k)}_{gconv}$ takes the feature vectors of a node's $k$-hop neighbors as input and outputs the weighted average of them.


The operator in (\ref{ho}) elegantly implements our idea of $k$-th order convolution on a graph, which is the convolution with kernel size of $k$ in conventional terminologies of CNN. On one hand, it can be viewed as an efficient high-order generalization of the first-order graph convolution in (\ref{node-GCN}). On the other hand, this operator is closely related to the graph spectral convolution in (\ref{spectral}), as the $k$-th order polynomial on the graph spectrum can also be regarded as an operation within the scope of $k$-hop neighborhood $\mathcal{N}_j$.

\subsection{Adaptive Filtering Module}
Based on the operator (\ref{ho}), we now introduce an adaptive filtering module for graph convolution. It filters the convolution weights according to the the features and the neighborhood connection of a specific node.  Take the molecule graph in chemistry for example, benzene rings are more important than alkyl chains when predicting the properties of molecules. As a result, we desire larger convolution weights for neighborhood atoms on the benzene rings than alkyl chains. Without the adaptive module, graph convolutions are spatially invariant and fails to work as desired. The introduction of adaptive filters will allow the network to find the convolution target adaptively and to better capture the locality disparities. 

The idea of the adaptive filtering comes from the attention mechanism \cite{xu2015show}, which chose the interest pixels adaptively while generating the corresponding words in the output sequence. It can also be viewed as a variant of the gates that optionally let information through in Long Short-Term Memory (LSTM) network \cite{hochreiter1997long}. Technically, our adaptive filter is a nonlinear operator $g$ on the weight matrix $W_k$, i.e. 
\begin{equation}
\widetilde{W_k} = g \circ W_k,
\label{ad}
\end{equation}
where $\circ$ denotes element-wise matrix product. In fact, the operator $g$ is determined together by $\widetilde{A}^k$ and $X$, reflecting both node features and graph connections,
\begin{equation*}
g=f_{adp}\left(\widetilde{A}^k,X\right).
\end{equation*}

We consider two candidates for the function $f_{adp}$:
\begin{equation}
f_{adp/prod} = \mathrm{sigmoid}\left(\widetilde{A}^k X Q\right)
\label{eq_adp_prod}
\end{equation}

and
\begin{equation}
f_{adp/lin} = \mathrm{sigmoid}\left(Q\cdot \left[\widetilde{A}^k,X\right]\right).
\label{eq_adp_lin}
\end{equation}

Here and hereafter, $[\cdot,\cdot]$ refers to the matrix concatenation. The first operator considers the interaction of node features and graph connections via an inner product for $A$ and $X$ while the second one does so via linear transformation.  In practice, we find that the linear adaptive filter (\ref{eq_adp_lin}) achieves a better performance than the product one (\ref{eq_adp_prod}) on almost all tasks. Therefore, we will adopt and report the performance based on the linear one in the experiment section. The adaptive filters are designed for a weighted selection of nodes, therefore a sigmoid non-linearity is applied to binarize its values. The parameter matrix $Q$ will align the output dimension of $f_{adp}$ to be the same with matrix $A$. Unlike the existing design of dynamic filters which generate the weights solely from node or edge features, our adaptive filtering module provides a more thorough consideration by taking both node features and graph connections into account.

\subsection{The Framework of HA-GCN}

In this subsection, we present the framework of HA-GCN and demonstrate how it can be applied to various tasks. By adding the adaptive module (\ref{ad}) into the high-order convolution operator (\ref{ho}), we define the HA operator:
\begin{equation*}
\widetilde{L}^{(k)}_{HA} = \left(\widetilde{W}_k \circ \widetilde{A}^k\right)X+B_k,
\end{equation*}

\begin{figure}[ht!]
\centering
\begin{tabular}{ll}
 a)& b)\\
 \includegraphics[width=.48\linewidth]{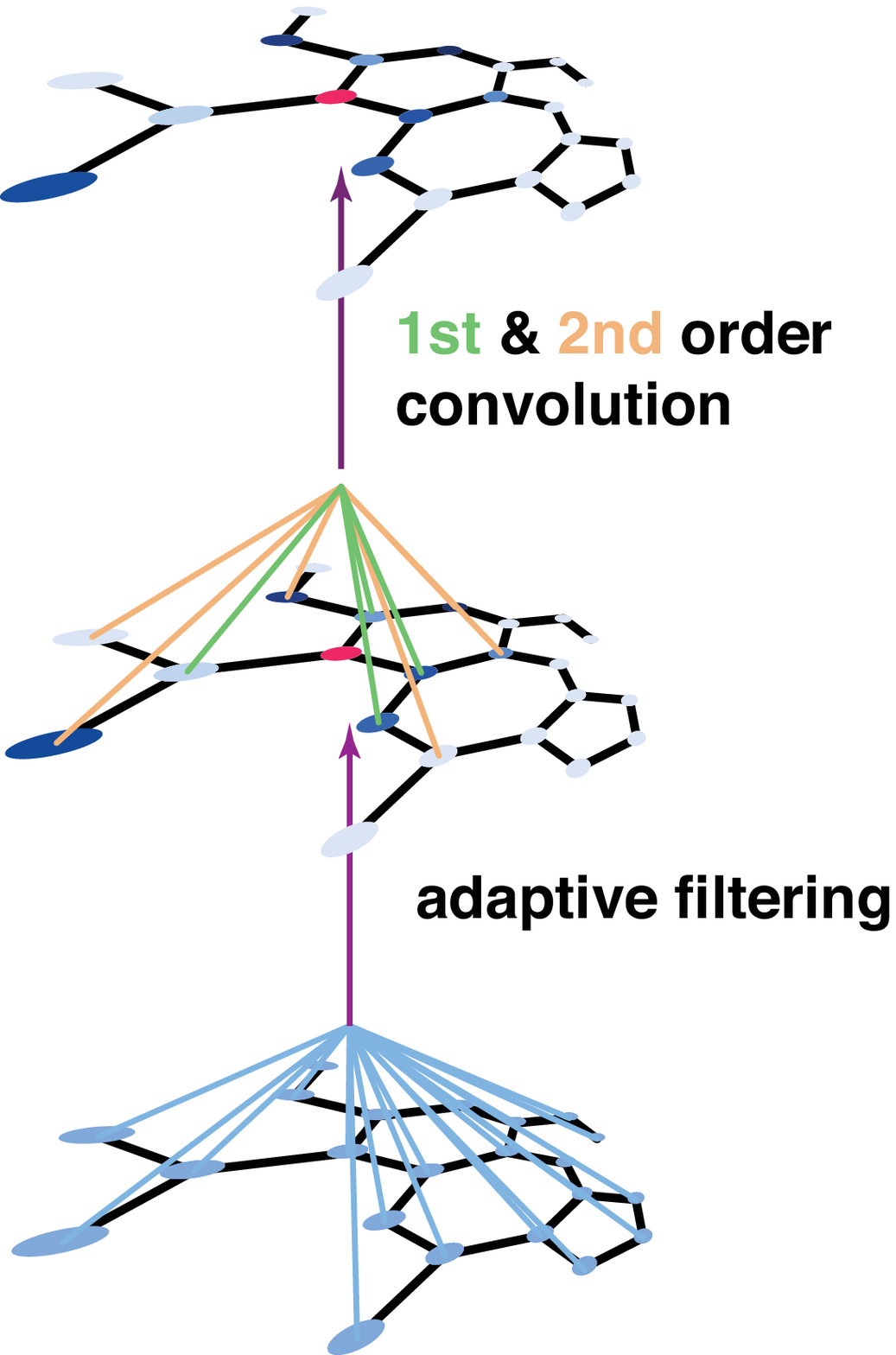} & \includegraphics[width=.41\linewidth]{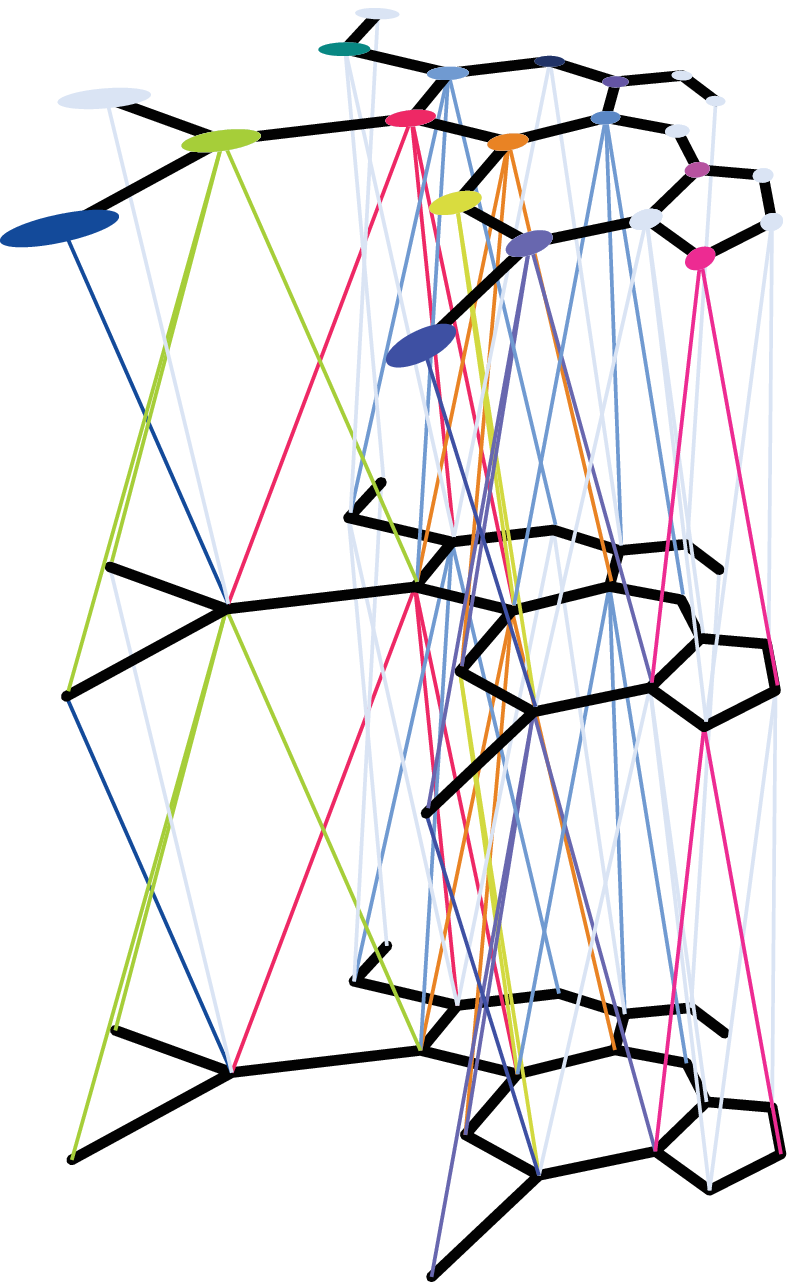}\\
\end{tabular}
\caption{The illustration of (a) high-order convolution and adaptive filtering module and (b) the whole graph convolutional network.}
\label{illus}
\end{figure}

Figure \ref{illus} gives a visualization of the operators and the framework HA-GCN. Figure \ref{illus}(a) illustrates the operator $\widetilde{L}^{(k)}_{HA}$ for a single node with $k=2$: the bottom layer of adaptive filter $g$ applies to weight matrices $W_1$ and $W_2$ to obtain the adaptive weights $\widetilde{W}_1$ and $\widetilde{W}_2$ (illustrated by the orange and green lines); the second layer brings the adaptive weights and the corresponding adjacency matrix together for convolution. Figure \ref{illus}(b) emphasizes the fact that the convolution is operated on each node in the graph, with a layer-by-layer manner. It is important to notice that the high-order operator and adaptive filtering module (HA operators) can be used together with other neural network architectures/operations like fully-connected layers, pooling layers, and non-linear transformations. In this paper, we name the graph convolutional network architecture built with our HA operator as HA-GCN.

After all layers of convolution, the features from different orders of convolution are concatenated together:
\begin{equation*}
L_{HA} = \left[\widetilde{L}^{(1)}_{HA},... ,\widetilde{L}^{(K)}_{HA}\right].
\end{equation*}

The framework of HA-GCN takes a feature matrix of $X \in \mathcal{R}^{n\times m}$ ($n$ is the number of nodes in the network and $m$ is the dimension of node features) and outputs a matrix of shape $n\times (mK)$, resulting in an increase of the feature dimension by a factor of $K$. Now, we elaborate more on how to apply HA-GCN on various tasks.

\textbf{Node-centric prediction:} After the graph convolutions in HA-GCN, each node is associated with a feature vector. The feature vectors can be used for tasks of node-centric classification or regression. It is also closely related to the graph (network) representation learning \cite{perozzi2014deepwalk,grover2016node2vec}, which refers to the procedure of learning a feature vector for each individual in a complicated system. Under node-centric settings, it means to learn a vector for each node in the graph that meaningfully reflects the local graph structure around that node. Our HA-GCN also outputs a vector for each node in the graph. In this sense, HA-GCN can be viewed as a supervised graph representation learning framework.

\textbf{Graph-centric prediction:} To handle graphs of different sizes, the input adjacency matrix and feature matrix are padded with zero on the bottom right. Here we point out a subtle difference between node-centric and graph-centric tasks: Under node-centric settings, the dataset is a single network with part of the nodes' label/value used as training set and the others as validation and test set, while under graph-centric settings, the dataset is a set of graphs (possibly of different sizes), divided into training/validation/test set. The HA-GCN works for both cases and the number of parameters in HA convolutional layer is $O(n^2)$ with $n$ being the size of the graph (or the maximum size of the graphs). As will be demonstrated later in the experiment section, HA-GCN is more prone to over-fitting under node-centric settings than graph-centric settings.

\textbf{Graph generative modeling:} The task of graph generative modeling refers to the learning of a probabilistic model from a set of graphs $\bar{\mathcal{G}} = \{\mathcal{G}_1,...,\mathcal{G}_N\}$, with which we can sample graphs that are unseen before but still have similar structures with the graphs in $\bar{\mathcal{G}}$. With the adventure of variational auto-encoder \cite{kingma2013auto} and adversarial auto-encoder \cite{makhzani2015adversarial}, graph convolutional networks can be made suitable for the task of generative modeling in addition to discriminative modeling. 

An auto-encoder always consists of two parts: an encoder and a decoder. The encoder maps the input data $X \in \mathcal{X}$ to an encoding vector $Y \in \mathcal{Y}$ and the decoder maps from $\mathcal{Y}$ back to $\mathcal{X}$. We call the encoding space $\mathcal{Y}$ latent or hidden space. To make it a generative model, we usually assume a probabilistic distribution (for example a Gaussian distribution) over the latent space. Here we consider the usage of HA-GCN as encoder for graph generative modeling. Given the length of the paper, we skip the technical discussion of the auto-encoder model here and defer more details about the HA-GCN auto-encoder architecture to the experiment section. As an application, the graph generative models allow us to create a continuous representation of molecules and generate new chemical structures by searching the latent space,  which can be used to guide the process of material design or drug screening.

\section{Experiments}

\subsection{Node-centric learning}

First, we considered a node-centric task of supervised document classification in citation graphs. The datasets \cite{Sen:2008wi} have three citation graphs, where each graph contains bag-of-words feature vectors for every document and a list of citation links between documents. We treated the citation links as (undirected) edges and construct a binary and symmetric adjacency matrix $A$. Each document has a class label and the goal is to predict the class label from the document feature and the citation graph. The statistics of the datasets are as reported in \cite{Sen:2008wi}.
\begin{table}[ht!]
\begin{center}
  \resizebox{0.45\textwidth}{!}{
\begin{tabular}{ccccc} 
\toprule
Dataset & Nodes & Edges & Classes &  Features \\ \hline  
\rule{0pt}{2.5ex}
Citeseer &3,327& 4,732& 6& 3,703 \\
Cora &2,708& 5,429&7&1,433 \\
Pubmed &19,717&44,338&210&5,414 \\
\bottomrule 
\end{tabular}}
\end{center}
\end{table}

\textbf{Training and Architecture}: We used the same GCN network structure of Kipf et al. \cite{kipf2016semi}, except a replacement of their first-order graph convolutional layer with our HA layer. Here and hereafter, we use \texttt{gcn\_\{$1,...,k$\}}
to denote graph convolutional layer up with order $1,\cdots, k$. 
\texttt{fc}$k$ refers fully connected layer with $k$ hidden units.

\begin{center}
  \resizebox{0.45\textwidth}{!}{
\begin{tabular}{c c} 
\toprule
Name & Architectures \\ \hline  
\rule{0pt}{2.5ex}
GCN & gcn\_\{1\}-fc128-gcn\_\{1\}-fc1-softmax\\
\texttt{gcn\_\{1, 2\}} & gcn\{1,2\}-fc128-gcn\{1,2\}-fc1-softmax\\
\texttt{adp\_gcn\_\{1, 2\}} & adp\_gcn\{1,2\}-fc128-adp\_gcn\{1,2\}-fc1-softmax\\
\bottomrule 
\end{tabular}}
\end{center}

\begin{table}[ht]
\begin{center}
  \resizebox{0.45\textwidth}{!}{
\begin{tabular}{cccc} 
\toprule
Method & Citeseer & Cora & Pubmed  \\ \hline  
\rule{0pt}{2.5ex}
l1\_logistic & 0.653 & 0.701 & 0.693 \\
l2\_logistic & 0.672 & 0.724 & 0.685 \\
DeepWalk & 0.631 & 0.746 & 0.712 \\
Planetoid & 0.724 & 0.832 & 0.844 \\
GCN & 0.776 & 0.889 & 0.839 \\
\texttt{gcn\_\{1,2\}} & \textbf{0.788} & \textbf{0.901} & \textbf{0.851}\tabularnewline
\texttt{adp\_gcn\_\{1,2\}} & 0.765 & 0.862 & 0.840\\
\bottomrule 
\end{tabular}
}
\caption{\small The accuracy of node classification results. l1\_logistic and l2\_logistic stand for $l1$ and $l2$ regularized logistic regression, DeepWalk refers to the algorithm by Perozzi et al. \cite{perozzi2014deepwalk}, Planetoid refers to the algorithm by Yang et al. \cite{Yang:2016ts}, and GCN refers to the graph convolutional neural network by Kipf et al. \cite{kipf2016semi}. All the models are implemented with the open-source code on github.}
\label{nodetask}
\end{center}
\end{table}

 To compare the performance of different models, we randomly divided the dataset into training/validation/test sets with a ratio of $7:1.5:1.5$ and reported the prediction accuracy on test set in Table \ref{nodetask}. The hyper-parameters are $0.7$  (dropout rate), $0.5\cdot 10^{-8}$ (L2 regularization), and $128$ (hidden units). From the perspective of node representation learning, the first three models are unsupervised but the last four are (semi-)supervised. This explains why the later ones have better performance. With our second-order HA graph convolution, the information from $2$-hop neighbors can be utilized, resulting in an approximately $2$\% increment of accuracy. Also, the adaptive module fails to further improve the accuracy. This is because the adaptive filter is designed to generate different filter weights for different graphs. However, each node-centric task has only one graph, whose convolution weights can be learned directly. Therefore, the adaptive module becomes redundant in this node-centric setting.

\subsection{Graph-centric learning}

In this experiment, we demonstrated the performance of HA-GCN on prediction tasks for molecule graphs. The goal is to predict the molecule's properties based on a molecule graph. We used the same datasets as described in Duvenaud et al. \cite{duvenaud2015convolutional} and evaluate the following three properties:
\begin{itemize}
\item Solubility: The aqueous solubility of 1144 molecules by \cite{delaney2004esol}. 
\item Drug efficacy: The half-maximal effective concentration (EC50) in
vitro of 10,000 molecules against a sulfide-resistant strain of P.
falciparum, the parasite that causes malaria, as measured by \cite{gamo2010thousands}. 
\item Organic photovoltaic efficiency: The Harvard Clean Energy Project
\cite{hachmann2011harvard} uses expensive DFT simulations to estimate
the photovoltaic efficiency of 30,000 organic molecules. 
\end{itemize}
With the same process described in Duvenaud et al. \cite{duvenaud2015convolutional}, we first used RDKit \cite{landrum2006rdkit} to convert the SMILE \cite{weininger1988smiles} representation of molecules into graphs, which treats hydrogen atoms implicitly. Each node in the graph corresponds to an atom and is appended with a $d$-dimensional initial feature vector. The features concatenate a one-hot encoding of the atoms element, its degree, the number of attached hydrogen atoms, and the implicit valence, and an aromaticity indicator.

\textbf{Training and Architecture}: The following network architectures are used for comparison. \texttt{l1\_gcn} and \texttt{l2\_gcn} refer to convolutional networks with $1$ and $2$ graph convolutional layer(s), respectively.  To compare the performance of different models, we reported the root of mean square errors (RMSEs) in Table \ref{graphtask}.
\begin{center}
  \resizebox{0.45\textwidth}{!}{
\begin{tabular}{c c} 
\toprule
Name & Architectures \\ \hline  
\rule{0pt}{2.5ex}
\texttt{l1\_gcn} & gcn\_\{1,2,3\}-ReLU-fc64-ReLU-fc16-ReLU-fc1\\
\texttt{l1\_adp\_gcn} & adp\_gcn\{1,2,3\}-ReLU-fc64-ReLU-fc16-ReLU-fc1\\
\texttt{l2\_gcn} & {[}gcn\_\{1,2,3\}-ReLU{]}{*}2-fc64-ReLU-fc16-ReLU-fc1\\
\texttt{l2\_adp\_gcn} & {[}adp\_gcn\_\{1,2,3\}-ReLU{]}{*}2-fc64-ReLU-fc16-ReLU-fc1\\
\bottomrule 
\end{tabular}}
\end{center}

\begin{table}[ht!]
\begin{center}
  \resizebox{0.45\textwidth}{!}{
\begin{tabular}{c c c c }
\toprule
\multirow{3}{*}{Model} & \multicolumn{3}{c}{Dataset} \vspace{0.1cm}
\\
\cline{2-4} 
\rule{0pt}{2.5ex}
 & \multirow{2}{*}{Solubility} & Drug  & Photovoltaic\\
 &  & efficacy & efficiency \vspace{0.1cm} \tabularnewline
\cline{1-4} 
\rule{0pt}{2.5ex}
NFP  & 0.52 & 1.16 & 1.43\\
MGC  & 0.46 & 1.07 & 1.10\\
node-GCN & 0.54 & 1.14 & 1.45\\
\texttt{l1\_gcn} & 0.61 & 1.20 & 1.54\\
\texttt{l1\_adp\_gcn} & 0.50 & 1.17 & 1.24\\
\texttt{l2\_gcn} & 0.56 & 1.09 & 1.35\\
\texttt{l2\_adp\_gcn} & \textbf{0.38} & \textbf{1.07} & \textbf{1.08}\\

\bottomrule 
\end{tabular}
}
\caption{\small{Prediction RMSEs: NFP refers to neural fingerprint \cite{duvenaud2015convolutional} and MGC refers to molecular graph convolution \cite{kearnes2016molecular}. Their performances are taken from the original papers.  The node-GCN refers to the graph convolutional network \cite{kipf2016semi} and is implemented with the open-source code provided by the authors.}}
\label{graphtask}
\end{center}
\end{table}

The model node-GCN is indeed a first-order HA-GCN without adaptive filtering module. From the comparison between node-GCN, \texttt{l1\_gcn} and \texttt{l2\_gcn}, we can see the effectiveness of our high-order convolution operator. Also, the networks with adaptive modules have a uniformly better performance than their counterparts without the module, which demonstrates its advantage. 

\subsection{Graph Generative Modeling}

In this experiment, we considered the task of graph generative modeling with HA-GCN auto-encoder. The network architectures are

\begin{center}
  \resizebox{0.45\textwidth}{!}{
\begin{tabular}{c c} 
\toprule
Name & Architectures \\ \hline  
\rule{0pt}{2.5ex}
\texttt{gcn\_encoder} & gcn\_\{1,2,3\}-ReLU-fc64-ReLU-fc16\\
\texttt{gcn\_decoder} & fc16-fc64-dconv-ReLU\\
\bottomrule 
\end{tabular}}
\end{center}

where \texttt{gcn}\_\{$k$\} and \texttt{fc} are defined as before, and \texttt{dconv} is defined as $L_{dconv}=\sigma\left(AA^{\top}\right)$.

We implemented HA-GCN as the encoder for both variational auto-encoder (VAE) and adversarial auto-encoder (AAE). As stated before, the graph generative models can be used to guide the molecule synthesis and the model performance is evaluated based on the proportion of valid molecules in all the newly-sampled molecules. We compared our HA-GCN generative model with the state-of-the-art RNN model of Grammar Variational Autoencoder (RNN-GVAE) \cite{Kusner:2017tv} We closely followed the experiment setup as in the graph variational autoencoder by Kipf et al. \cite{Kipf:2016ul}, with
a training data of $250,000$ SMILES molecules
\cite{weininger1988smiles} extracted randomly from the ZINC database by Gómez-Bombarelli et al. \cite{GomezBombarelli:2016vk}. Then  $10,000$ encodings were drawn from a normal distribution in the latent space, and decoded to generate molecules. The HA-GCN-AAE model got $23.6$\% valid molecules, and HA-GCN-VAE got $20.3$\%, while the RNN-GVAE got $15.3$\%. Here we achieved a significant gain in performance with HA-GCN as the encoder for graph generative modeling.

\section{Visualization of the HA-GCN}

\subsection{Visualization of the Convolution Weights}

\begin{figure}[ht!]
  \includegraphics[width=.45\textwidth]{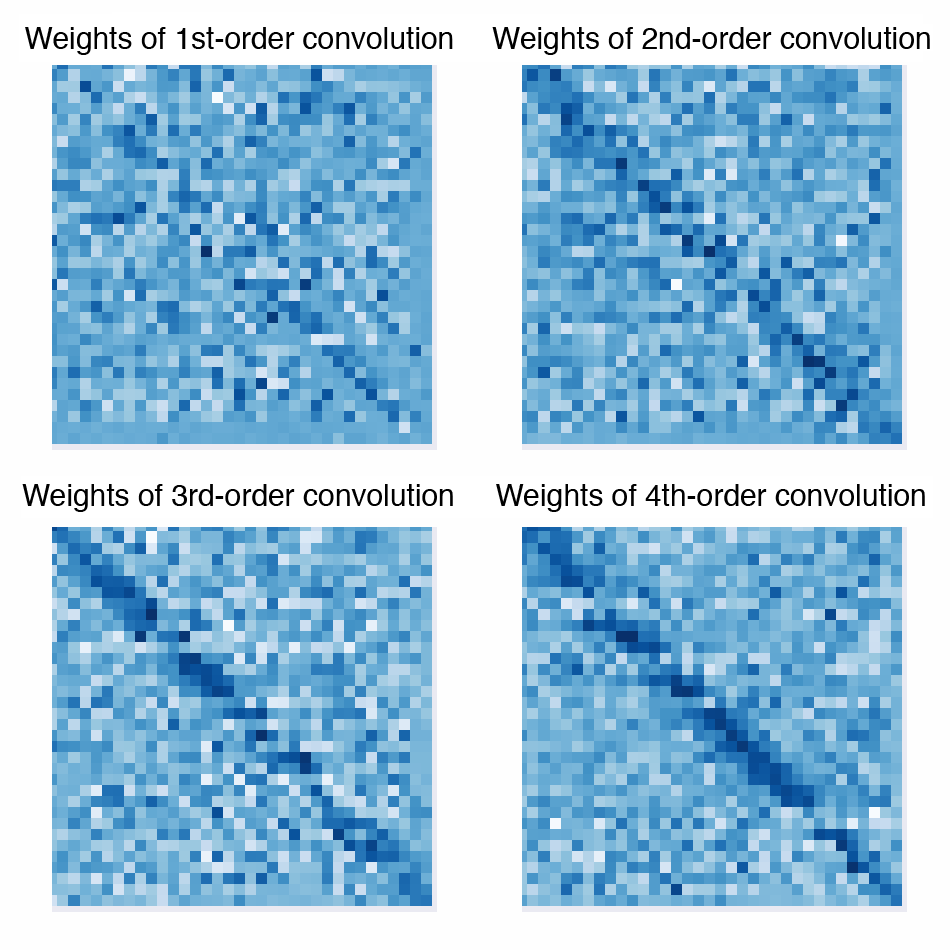}
\captionof{figure}{Visualization of the convolution weights.}
 \label{weight}
\end{figure}

We visualized the convolution weights $W_k$ in (\ref{ho}) in the HA-GCN trained for the task of photovoltaic efficiency prediction. The convolutional weight matrices are plotted as in Figure \ref{weight} and the darkness of a block corresponds to the weight value of corresponding node. We have following observations from the plots: First, it is easy to see that the weight matrices have symmetrical patterns, which is due to the symmetry of adjacency matrix $A$ (or $\widetilde{A}$). Second, as the order of convolution increases, the weights on the central nodes increase as well. Our explanation is that as the order of convolution increases, there are more nodes in the reception field. The weight increments on the central nodes is to balance off the effect of having more nodes within the scope of convolution. Third, for weights of convolution orders larger than $1$, we observe many off-diagonal blocks having large values (with dark blue color), showing the necessity of introducing the high-order convolution.

\begin{figure*}[t!]
  \centering
  \includegraphics[width=0.9\textwidth]{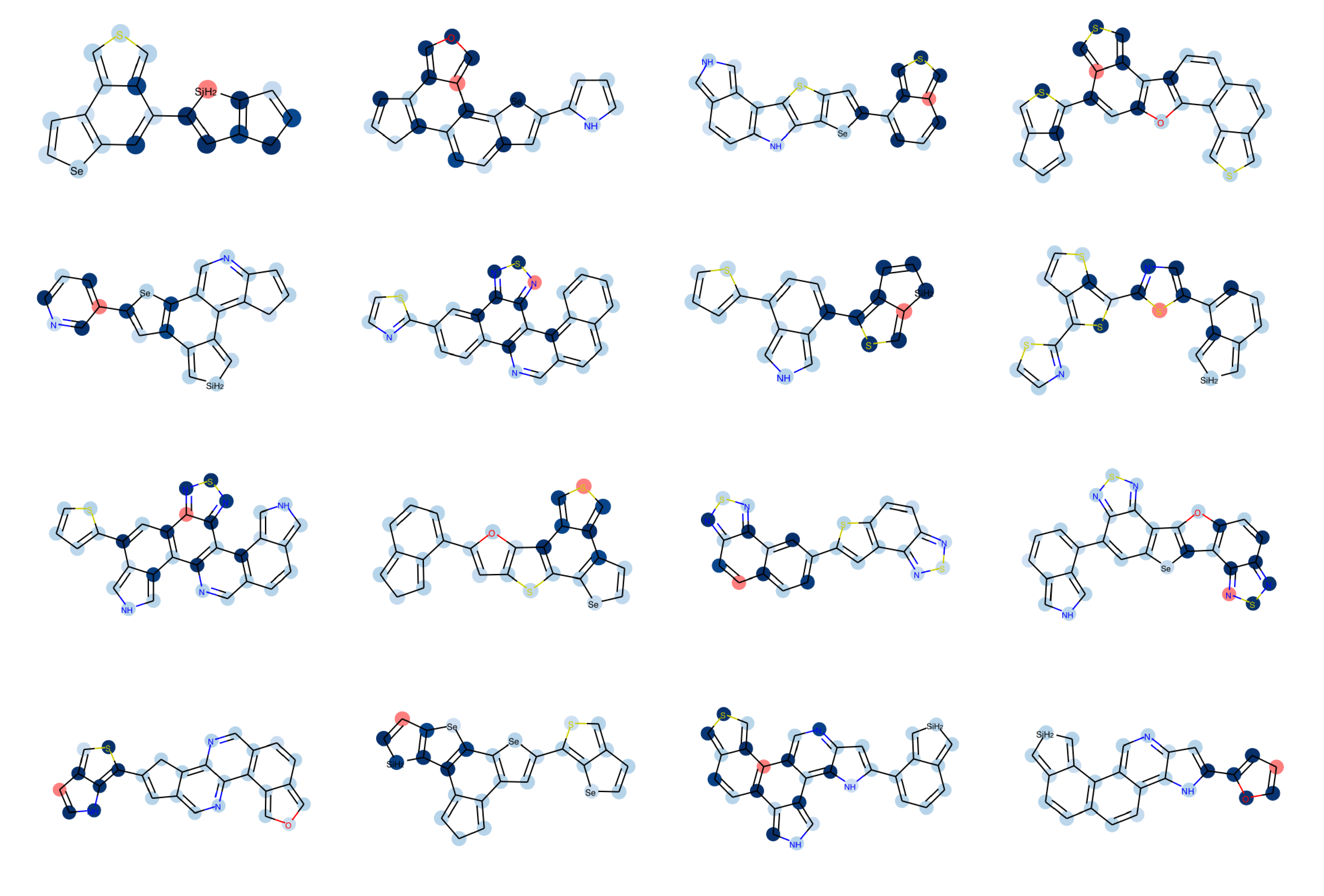}
\caption{Visualization of the filter weights. The atoms highlighted with red is the randomly selected central node for convolution, the blue color on the atoms indicate the filter weights, with darker blue meaning larger weight.}
\label{adapt}
\end{figure*}

\subsection{Visualization of Adaptive Filters}

The adaptive filters $g$ in (\ref{ad}) learned from graph connections and node features are visualized in Figure \ref{adapt}. The atoms highlighted in red are the randomly selected central nodes for convolution, and the blue circles on atoms indicate the filter weights, with darker blue meaning larger weight. We have following observations: First, the adaptive filter weights are almost binarized, which means that the filters are capable of selecting nodes for convolution adaptively based on the features and connectivity. Second, for almost all molecules in Figure \ref{adapt}, the atoms being selected are atoms on aromatic rings, which agrees with the chemical intuition that aromatic rings are more important than alkyl chains in terms of predicting organic photovoltaic efficiency. Another interesting observation is that the adaptive filter automatically learned the ortho-para rule in chemistry, which states that for the benzene ring, the functional groups on the opposite of (ortho) and next to (para) a specific atom has a greater influence on the properties of that atom than the functional groups on other sites. For example, in the molecule on Figure \ref{adapt} row 2 column 1, the  weights on the atoms which are opposite of and next to the central atom are selected against other atoms on the six-member ring.

\section{Conclusion}

In this work, we developed a graph convolutional network architecture of HA-GCN with two new convolution modules specially designed for graph-structured data. With experiments showing the effectiveness of those modules, we strongly advocate a consideration of them in all the graph convolutional network architecture design. For future works, on one hand, we believe that it still deserves more work on designing the convolution network with a careful thought on the underlying graph (global and local) structure. On the other hand, we are currently conducting experiments on automatic chemical design to further demonstrate the practical value of our framework. 

\bibliography{biblio}
\bibliographystyle{aaai}

\end{document}